\documentclass{article}

\usepackage{times}
\usepackage{graphicx} 
\usepackage{subfigure} 

\usepackage{natbib}
\usepackage{bm}

\usepackage{algorithm}
\usepackage{color}
\usepackage{pbox}

\usepackage{algorithmic}
\usepackage{hyperref}

\usepackage[accepted]{icml2016} 

\usepackage{url} 
\usepackage{bm}
\usepackage{bbm}
\usepackage{amsmath,amssymb}
\usepackage{xspace}
\usepackage{bm} 
\usepackage{dsfont}
\usepackage{psfrag}
\usepackage{rotating}
\usepackage{epstopdf}
\usepackage{booktabs}

\floatstyle{ruled}
\newfloat{algo}{thp}{lop} 
\floatname{algo}{Algorithm}

\newcommand{\OmegaSpace}{\Omega}

\newcommand{\diag}{\mathop{\textrm{diag}}}

\newcommand{\transp}{^{T}}

\newcommand{\proba}{P}
\newcommand{\grad}{\nabla}

\newcommand{\U}{\bm{U}}

\newcommand{\complexSpace}{\mathbb{C}}
\renewcommand{\Re}{\mathbb{R}}
\newcommand{\C}{\complexSpace} 
\newcommand{\R}{\Re} 
\newcommand{\real}{\mathrm{Re}}
\newcommand{\imag}{\mathrm{Im}}
\newcommand{\randn}{\mathrm{randn}}
\newcommand{\srank}{\mathrm{rank}_{\pm}}
\newcommand{\lrank}{\mathrm{rank}}
\newcommand{\sign}{\mathrm{sign}}

\providecommand{\U}[1]{\protect\rule{.1in}{.1in}}
\newcommand{\be}{\begin{equation}}
\newcommand{\ee}{\end{equation}}
\newcommand{\bd}{\begin{definition}}
\newcommand{\ed}{\end{definition}}
\newcommand{\ba}{\begin{algorithm}}
\newcommand{\ea}{\end{algorithm}}
\newcommand{\br}{\begin{problem}}
\newcommand{\er}{\end{problem}}
\newcommand{\bex}{\begin{example}}
\newcommand{\eex}{\end{example}}
\newcommand{\bt}{\begin{theorem}}
\newcommand{\et}{\end{theorem}}

\newtheorem{theorem}{Theorem}

\newtheorem{definition}[theorem]{Definition}
\newtheorem{example}[theorem]{Example}

\newtheorem{problem}[theorem]{Problem}

\def\ifa{\iffalse}
\def\ifappendix{\iftrue}

\newcommand{\Relation}{\mathbf{X}}
\newcommand{\ObsTensor}{\mathbf{Y}}
\newcommand{\EntitySpace}{\mathcal{E}}
\newcommand{\RelationSpace}{\mathcal{R}}

\newcommand{\rank}{K} 
\newcommand{\setent}{\mathcal{E}} 
\newcommand{\setrel}{\mathcal{R}} 
\newcommand{\eemb}{e} 
\newcommand{\Eemb}{E} 
\newcommand{\wemb}{w} 

\def\tt{\texttt}

\usepackage[textsize=tiny]{todonotes}

\icmltitlerunning{Complex Embeddings for Simple Link Prediction}

\begin{document} 

\twocolumn[
\icmltitle{Complex Embeddings for Simple Link Prediction}

\icmlauthor{Th\'eo Trouillon$^{1,2}$}{theo.trouillon@xrce.xerox.com}
\icmlauthor{Johannes Welbl$^{3}$}{j.welbl@cs.ucl.ac.uk}
\icmlauthor{Sebastian Riedel$^{3}$}{s.riedel@cs.ucl.ac.uk}
\icmlauthor{\'Eric Gaussier$^{2}$}{eric.gaussier@imag.fr}
\icmlauthor{Guillaume Bouchard$^{3}$}{g.bouchard@cs.ucl.ac.uk}
\icmladdress{$^{1}$ Xerox Research Centre Europe, 6 chemin de Maupertuis, 38240 Meylan, FRANCE\\$^{2}$ Universit\'e Grenoble Alpes, 621 avenue Centrale, 38400 Saint Martin d'H\`eres, FRANCE\\$^{3}$ University College London, Gower St, London WC1E 6BT, UNITED KINGDOM}

\icmlkeywords{embeddings, factorization, knowledge graph, complex numbers, link prediction, statistical relational learning, complex embeddings, }

\todototoc

\vskip 0.3in
]

\begin{abstract} 
In statistical relational learning, the link prediction problem is key to automatically understand the structure of large knowledge bases. As in previous studies, we propose to solve this problem through latent factorization. However, here we make use of complex valued embeddings. The composition of complex embeddings can handle a large variety of binary relations, among them symmetric and antisymmetric relations. Compared to state-of-the-art models such as Neural Tensor Network and Holographic Embeddings, our approach based on \emph{complex} embeddings is arguably \emph{simpler}, as it only uses the Hermitian dot product, the complex counterpart of the standard dot product between real vectors. Our approach is scalable to large datasets as it remains linear in both space and time, while consistently outperforming alternative approaches on standard link prediction benchmarks.\footnote{Code is available at: \url{https://github.com/ttrouill/complex}}
\end{abstract}

\section{Introduction}

Web-scale knowledge bases (KBs) provide a structured representation of world knowledge, with projects such as DBPedia~\cite{dbpedia}, Freebase~\cite{Bollacker2008} or the Google Knowledge Vault~\cite{Dong:2014:KnowledgeVault}. 
They enable a wide range of applications such as recommender systems, question answering or automated personal agents. The incompleteness of these KBs has stimulated \mbox{research} into predicting missing entries, a task known as link prediction that is one of the main problems in Statistical Relational Learning~\citep[SRL,][]{Getoor2007}.




KBs express data as a directed graph with labeled edges (relations) between nodes (entities). Natural redundancies among the recorded relations often make it possible to fill in the missing entries of a KB. 
As an example, the relation \tt{CountryOfBirth} is not recorded for all entities, 
but it can easily be inferred if the relation \tt{CityOfBirth} is known. 
The goal of link prediction is the automatic discovery of such regularities. However, many relations are non-deterministic:  the combination of the two facts \tt{IsBornIn(John,Athens)} and \tt{IsLocatedIn(Athens,Greece)} does not always imply the fact \tt{HasNationality(John,Greece)}. 
Hence, it is required to handle other facts involving these relations or entities in a probabilistic fashion.

To do so, an increasingly popular method is to state the link prediction task as a 3D binary tensor completion problem, where each slice is the adjacency matrix of one relation type in the knowledge graph.
Completion based on low-rank factorization or \emph{embeddings} has been popularized with the Netflix challenge \cite{koren_netflix}. A partially observed matrix or tensor is decomposed into a product of embedding matrices with much smaller rank, resulting in fixed-dimensional vector representations 
for each entity and relation in the database. 
For a given fact \emph{r(s,o)} in which subject $s$ is linked to object $o$ through relation $r$, the score can then be recovered as a multi-linear product between the embedding vectors of $s$, $r$ and $o$ ~\cite{nickel_2016_review}.


Binary relations in KBs exhibit various types of patterns: hierarchies and compositions like \tt{FatherOf}, \tt{OlderThan} or \tt{IsPartOf}---with partial/total, strict/non-strict orders---and  equivalence relations like \tt{IsSimilarTo}. As described in \citet{Bordes2013}, a relational model should (a) be able to learn all \mbox{combinations} of these properties, namely reflexivity/irreflexivity, symmetry/antisymmetry and transitivity, and (b) be linear in both time and memory in order to scale to the size of present day KBs, and keep up with their growth. 

Dot products of embeddings scale well and can naturally handle both symmetry and (ir-)reflexivity of relations; using an appropriate loss function even enables transitivity~\cite{bouchard2015}.
However, dealing with antisymmetric relations has so far almost always implied an explosion of the number of parameters \cite{Nickel2011,socher2013reasoning} (see Table \ref{tab:scoring}), making models prone to overfitting. Finding the best ratio between expressiveness and parameter space size is the keystone of embedding models.

In this work we argue that the standard dot product between embeddings can be a very effective composition function, provided that one uses the right \emph{representation}. Instead of using embeddings containing real numbers we discuss and demonstrate the capabilities of complex embeddings. When using complex vectors, i.e. vectors with entries in $\complexSpace$,  the dot product is often called the \emph{Hermitian} (or sesquilinear) dot product, as it involves the conjugate-transpose of one of the two vectors. 
As a consequence, the dot product is not symmetric any more, and facts about antisymmetric relations can receive different scores depending on the ordering of the entities involved.
Thus complex vectors can effectively capture antisymmetric relations while retaining the efficiency benefits of the dot product, that is linearity in both space and time complexity. 


The remainder of the paper is organized as follows. We first justify the intuition of using complex embeddings in the square matrix case in which there is only a single relation between entities. The formulation is then extended to a stacked set of square matrices in a third-order tensor to represent multiple relations. We then describe experiments on large scale public benchmark KBs in which we empirically show that this representation leads not only to simpler and faster algorithms, but also gives a systematic accuracy improvement over current state-of-the-art alternatives.

To give a clear comparison with respect to existing approaches using only real numbers, we also present an equivalent reformulation of our model that involves only real embeddings. This should help practitioners when implementing our method, without requiring the use of complex numbers in their software implementation.

\section{Relations as Real Part of Low-Rank Normal Matrices}
In this section we discuss the use of complex embeddings for low-rank matrix factorization and illustrate this by \mbox{considering} a simplified link prediction task with merely a single relation type. 

Understanding the factorization in complex space leads to a better theoretical understanding of the class of matrices that can actually be approximated by dot products of embeddings. These are the so-called \emph{normal matrices} for which the left and  right embeddings share the same unitary basis.

\subsection{Modelling Relations}

Let $\EntitySpace$ be a set of entities with $|\EntitySpace|=n$. A relation between two entities is represented as a binary value $Y_{so}\in\{-1,1\}$, where $s\in\EntitySpace$ is the subject of the relation and $o\in\EntitySpace$ its object. Its probability is given by the logistic inverse link function:
\begin{equation}
    \proba(Y_{so}=1) = \sigma(X_{so})
    \enspace
    \label{observation-model0}
\end{equation}
where $X\in\R^{n\times n}$ is a latent matrix of scores, and $Y$ the partially observed sign matrix.

Our goal is to find a 
generic structure for $X$ that leads to a flexible approximation of common relations in real world KBs. Standard matrix factorization approximates $X$ by 
a matrix product $UV\transp$, where $U$ and $V$ are two functionally independent $n\times K$ matrices, $K$ being the rank of the matrix. Within this formulation it is assumed that entities appearing as subjects are different from entities appearing as objects. This means that the same entity will have two different embedding vectors, depending on whether it appears as the subject or the object of a relation. 
This extensively studied type of model is closely related to the singular value decomposition (SVD) and fits well to the case where the matrix $X$ is rectangular. 
However, in many link prediction problems, the same entity can appear as both subject
\emph{and} object. It then seems natural to learn joint embeddings of the entities, 
which entails sharing the embeddings of the left and right factors, as proposed 
by several authors to solve the link prediction problem \cite{Nickel2011,bordes2013translating,Yang2015}. 

In order to use the same embedding for subjects and objects, researchers have generalised the notion of dot products to \emph{scoring functions}, also known as \emph{composition functions}, that combine embeddings in specific ways. 
We briefly recall several examples of scoring functions in Table~\ref{tab:scoring}, as well
as the extension proposed in this paper. 

\begin{table*}[]
    \centering
    \resizebox{2.03\columnwidth}{!}{%
    \begin{tabular}{|l|l|l|l|l|l|}
        \hline
        \textbf{Model} &
        \textbf{Scoring Function} &
        \textbf{Relation parameters} &
        \textbf{$\mathcal{O}_{time}$}&
        \textbf{$\mathcal{O}_{space}$}
        \\
        \hline
        RESCAL \cite{Nickel2011} &
        $e_s^T W_r e_o$ &
        $W_r\in\Re^{K^2}$&
        $\mathcal{O}(K^2)$& 
        $\mathcal{O}(K^2)$
        \\
        \hline
        TransE  \cite{bordes2013translating}&
        $||(e_s + w_r) - e_o||_p$ &
        $w_r\in\Re^K$ &
        $\mathcal{O}(K)$& 
        $\mathcal{O}(K)$
        \\
        \hline
        NTN  \cite{socher2013reasoning}&
        $u_r\transp f(e_s W_r^{[1..D]}e_o + V_r \begin{bmatrix} e_s 
        \\  e_o \end{bmatrix} + b_r)$ &
        \pbox{20cm}{$W_r\in\Re^{K^2 D}, b_r\in\Re^{K}$\\$V_r \in\Re^{2KD} ,u_r\in\Re^{K} $} &
        $\mathcal{O}(K^2D)$& 
        $\mathcal{O}(K^2D)$
        \\
        \hline
        DistMult \cite{Yang2015}&  
        $<w_r, e_s, e_o>$ &
        $w_r \in\Re^K$ &
        $\mathcal{O}(K)$& 
        $\mathcal{O}(K)$
        \\
        \hline
        HolE \cite{nickel_2016_holographic}&
        $w_r^T ( \mathcal{F}^{-1}[\overline{\mathcal{F}[e_s]} \odot \mathcal{F}[e_o]]))$
        &
        $w_r \in\Re^K$ &
        $\mathcal{O}(K\log K)$&
        $\mathcal{O}(K)$
        \\
        \hline
        ComplEx & 
        $\real(<w_r, e_s, \bar{e}_o>)$ &
        $w_r\in\complexSpace^{K}$&
        $\mathcal{O}(K)$& 
        $\mathcal{O}(K)$
        \\
        \hline
    \end{tabular}
    }
    \caption{
     Scoring functions of state-of-the-art latent factor models for a given fact $r(s,o)$, along with their relation parameters, time and space (memory) complexity. The embeddings $e_s$ and $e_o$ of subject $s$ and object $o$ are in $\R^K$ for each model, except for our model (ComplEx) where $e_s,e_o \in \C^K$.  $D$ is an additional latent dimension of the NTN model. 
    $\mathcal{F}$ and $\mathcal{F}^{-1}$ denote respectively the Fourier transform and its inverse, and $\odot$ is the element-wise product between two vectors.
    }
    \label{tab:scoring}
\end{table*}

\newcommand{\normalSpace}{\mathcal{H}}
Using the same embeddings for right and left factors boils down to Eigenvalue decomposition: 
\begin{equation}
X=EWE^{-1}\enspace.
\label{eig}
\end{equation}
It is often used to approximate real symmetric matrices
such as covariance matrices, kernel functions and distance or similarity matrices. In these cases all eigenvalues and eigenvectors live in the real space and $E$ is orthogonal: $E\transp=E^{-1}$. 
We are in this work however explicitly interested in problems where matrices --- and thus the \mbox{relations} they represent ---
can also be antisymmetric. In that case eigenvalue decomposition is not possible in the real space; there only exists a decomposition in the complex space where embeddings $x\in\C^K$ are composed of a real vector component $\real(x)$ and an imaginary vector component $\imag(x)$. With complex numbers, the dot product, also called the \emph{Hermitian} product, or \emph{sesquilinear} form, is defined as: 
\begin{equation}
    \left< u,v \right>:= \bar{u}\transp v
\end{equation} 
where $u$ and $v$ are complex-valued vectors, i.e. $u=\real(u) + i\imag(u)$ with $\real(u)\in\Re^K$ and $\imag(u)\in\Re^K$ corresponding to the real and imaginary parts of the vector $u\in\C^K$, and $i$ denoting the square root of $-1$. We see here that one crucial operation is to take the conjugate of the first vector: $\bar{u}=\real(u) - i \imag(u)$.
A simple way to justify the Hermitian product for composing complex vectors is that it provides a valid topological norm in the induced vectorial space. For example, $\bar{x}\transp x=0$ implies $x=0$ while this is not the case for the bilinear form $x \transp x$ as there are many complex vectors for which $x\transp x=0$. 

Even with complex eigenvectors $E\in\C^{n\times n}$, the inversion of $E$ in the eigendecomposition of Equation~(\ref{eig}) leads to computational issues. Fortunately, mathematicians defined an appropriate class of matrices that prevents us from inverting the eigenvector matrix: we consider the space of \emph{normal matrices}, i.e. the complex $n \times n$ matrices $X$, such that $X \bar{X}\transp$ = $\bar{X}\transp X$. The spectral theorem for normal matrices states that a matrix $X$ is normal if and only if it is unitarily diagonalizable:
\begin{eqnarray}
X = E W \bar{E}^T
\label{eqn:eigendec}
\end{eqnarray}
where $W\in\C^{n \times n}$ is the diagonal matrix of eigenvalues 
(with decreasing modulus) and $E\in\C^{n\times n}$ is a unitary matrix of 
eigenvectors, with $\bar{E}$ representing its complex conjugate. 

The set of purely real normal matrices includes all
 symmetric and antisymmetric sign matrices (useful to model hierarchical
relations such as \tt{IsOlder}), as well as 
all orthogonal matrices (including permutation matrices), and many other matrices that are useful to represent binary relations, such as assignment matrices which represent bipartite graphs. However, far from all matrices expressed as $E W \bar{E}^T$ are
purely real, and equation \ref{observation-model0} requires the scores $X$ to be purely real. So we simply keep only the real part of the
decomposition:
\begin{eqnarray}
X = \real(E W \bar{E}^T)\enspace.
\label{eqn:eigendecreal}
\end{eqnarray}

In fact, performing this projection on the real subspace allows the exact
decomposition of \emph{any} real square matrix $X$ and not only normal ones,
as shown by \citet{trouillon_unitdiag2016}.



Compared to the singular value decomposition, the eigenvalue decomposition has two key differences:
\begin{itemize}
    \item The eigenvalues are not necessarily positive or real;
    \item The factorization~(\ref{eqn:eigendecreal}) is useful as the rows of
    $E$ can be used as vectorial representations of the entities corresponding 
    to rows and columns of the relation matrix $\Relation$. Indeed, for a given entity, its subject embedding vector is the complex conjugate of its object embedding vector. 
\end{itemize}

\subsection{Low-Rank Decomposition}

In a link prediction problem, the relation matrix is unknown and the goal is to recover it entirely from noisy observations. To enable the model to be \emph{learnable}, i.e. to generalize to unobserved links, some regularity assumptions are needed. Since we deal with binary relations, we assume that they have low \emph{sign-rank}. The sign-rank of a sign matrix is the smallest rank of a real matrix that has the same sign-pattern as~$Y$:
\begin{eqnarray}
    \srank(Y) = \min_{A\in \R^{m \times n}} \{\lrank(A) | \sign(A) = Y \}\enspace.
\end{eqnarray}

This is theoretically justified by the fact that the sign-rank is a natural complexity measure of sign matrices \mbox{\cite{Linial2007}} and is linked to learnability \cite{alon2015sign}, and empirically confirmed by the wide success of factorization models~\cite{nickel_2016_review}. 

If the observation matrix $Y$ is low-sign-rank, then our
model can decompose it with a rank at most the double of
the sign-rank of $Y$.
That is, for any $Y\in\{-1,1\}^{n\times n}$, there always exists a matrix $X = \real(E W \bar{E}^T)$ with the same sign pattern $\sign(X)=Y$, where the rank of $E W \bar{E}^T$ is at most twice the sign-rank of $Y$ \cite{trouillon_unitdiag2016}.

Although twice sounds bad, this is actually a good upper bound. Indeed, the sign-rank is often \emph{much} lower than the rank
of $Y$. For example, the rank of the $n \times n$ identity matrix
$I$ is $n$, but $\srank(I)=3$ \cite{alon2015sign}. By permutation
of the columns $2j$ and $2j+1$, the $I$ matrix corresponds to the
relation \texttt{marriedTo}, a relation known to be hard to
factorize \cite{Nickel2014}. Yet our model can express it in rank 6, for any $n$.



By imposing a low-rank $K \ll n$ on $E W \bar{E}^T$, only the first $K$ values of
$\diag(W)$ are non-zero. So we can directly have $E \in \C^{n\times K}$ and
$W \in \C^{K\times K}$. Individual relation scores $X_{so}$ between entities $s$ and $o$ can be predicted through the following product of their embeddings $e_s, e_o\in\C^K$:
\begin{eqnarray}
    X_{so} &=& \real(e_s\transp W \bar{e}_o)\enspace.
    \label{eq:one_rel_model}
\end{eqnarray}

We summarize the above discussion in three points:

\begin{enumerate}
    \item Our factorization encompasses all possible binary relations.
    \item By construction, it accurately describes both symmetric and antisymmetric relations.
    \item Learnable relations can be efficiently approximated
by a simple low-rank factorization, using complex numbers to represent the latent factors.
\end{enumerate}

\section{Application to Binary Multi-Relational Data}

The previous section focused on modeling a single type of relation; 
we now extend this model to multiple types of relations. 
We do so by allocating an embedding $w_r$ to each relation $r$, and by sharing the entity embeddings across all relations.

Let $\RelationSpace$ and $\EntitySpace$ be the set of relations and entities present in the KB. We want to recover the matrices of scores $\Relation_r$ for all the relations $r \in\RelationSpace$.
Given two entities $s$ and $o$ $ \in \EntitySpace$, the log-odd of the probability that the fact \emph{r(s,o)} is true is:
\begin{equation}
    \proba(\ObsTensor_{rso}=1) = \sigma(\phi(r,s,o;\Theta))\enspace
    \label{observation-model}
\end{equation}
where $\phi$ is a scoring function that is typically based on a factorization of the observed relations and $\Theta$ denotes the parameters of the corresponding model. While $\Relation$ as a whole is unknown, we 
assume that we observe a set of true and false facts 
$\{\ObsTensor_{rso}\}_{r(s,o)\in \OmegaSpace} \in \{-1,1\}^{|\OmegaSpace|}$, corresponding to the partially observed adjacency matrices of different relations, where $\OmegaSpace\subset\RelationSpace\otimes\EntitySpace\otimes\EntitySpace$ 
is the set of observed triples. The goal is to find the probabilities of entries 
$\ObsTensor_{r's'o'}$ being true or false for a set of targeted unobserved 
triples $r'(s',o')\notin\OmegaSpace$.

Depending on the scoring function $\phi(s,r,o;\Theta)$ used to predict the entries of the
tensor $\Relation$, we obtain different models. Examples of scoring functions are given in Table~\ref{tab:scoring}. Our model scoring function is:
\begin{eqnarray}
    \phi(r,s,o;\Theta) &=& \real(<w_{r}, \eemb_s, \bar\eemb_o>)\\
    \label{eqn:complex-dot1}
    &=& \real(\sum_{k=1}^K w_{rk} \eemb_{sk} \bar\eemb_{ok})\\
    \label{eqn:complex-dot}
    &=& \left<\real(w_r),\real(e_s), \real(e_o)\right>\notag\\
    &&+ \left<\real(w_r),\imag(e_s), \imag(e_o)\right> \notag\\
    &&+ \left<\imag(w_r), \real(e_s),\imag(e_o)\right> \notag\\
    &&- \left<\imag(w_r),\imag(e_s),\real(e_o)\right>
    \label{eqn:sesquilinear-dot}
\end{eqnarray}
where $w_r\in \C^K$ is a complex vector . These equations provide two interesting views of the model:
\begin{itemize}
    \item \emph{Changing the representation}: Equation~(\ref{eqn:complex-dot1}) would correspond to DistMult with real embeddings, but handles asymmetry thanks to the complex conjugate of one of the embeddings%
    \footnote{Note that in Equation~(\ref{eqn:complex-dot1}) we used the standard componentwise multi-linear dot product $<a,b,c> := \sum_k a_kb_kc_k$. This is not 
    the Hermitian extension as it is not properly defined in the linear algebra literature.}.
    \item  \emph{Changing the scoring function}: Equation~(\ref{eqn:sesquilinear-dot}) only involves real vectors corresponding to the real and imaginary parts of the embeddings and relations.%
\end{itemize}

One can easily check that this function is antisymmetric when $w_r$ is purely imaginary (i.e. its real part is zero), and symmetric when $w_r$ is real. 
Interestingly, by separating the real and imaginary part of the relation embedding $\wemb_r$,
we obtain a decomposition of the relation matrix $\Relation_r$ as the sum of a symmetric matrix
$\real( \Eemb\diag(\real(\wemb_r)) \bar\Eemb\transp )$
and a antisymmetric matrix
$\imag( \Eemb \diag(-\imag(\wemb_r)) \bar\Eemb\transp )$. 
Relation embeddings naturally act as weights on each latent dimension: $\real(w_r)$ over the symmetric, real part of $\left< e_o, e_s \right>$, and $\imag(w)$ over the antisymmetric, imaginary part
of $\left< e_o, e_s \right>$. Indeed, one has $\left< e_o, e_s \right> = \overline{\left< e_s, e_o \right>}$, meaning that $\real(\left< e_o, e_s \right>)$ is symmetric, while $\imag(\left< e_o, e_s \right>)$ is antisymmetric. This enables us to accurately describe both symmetric and antisymmetric relations between pairs of entities, while still using joint representations of entities, whether they appear as subject or object of relations. 

Geometrically, each relation embedding $w_r$ is an anisotropic
scaling of the basis defined by the entity embeddings $E$, followed by a projection
onto the real subspace.

\section{Experiments}
\label{sec:expe}

In order to evaluate our proposal, we conducted experiments on both synthetic and real datasets. The synthetic dataset is based on relations that are either symmetric or antisymmetric, whereas the real datasets comprise different types of relations found in different, standard KBs. We refer to our model as ComplEx, for Complex Embeddings.

\begin{figure*}{{\extracolsep{8pt}}}
	\centering
	\includegraphics[width=0.40\textwidth]{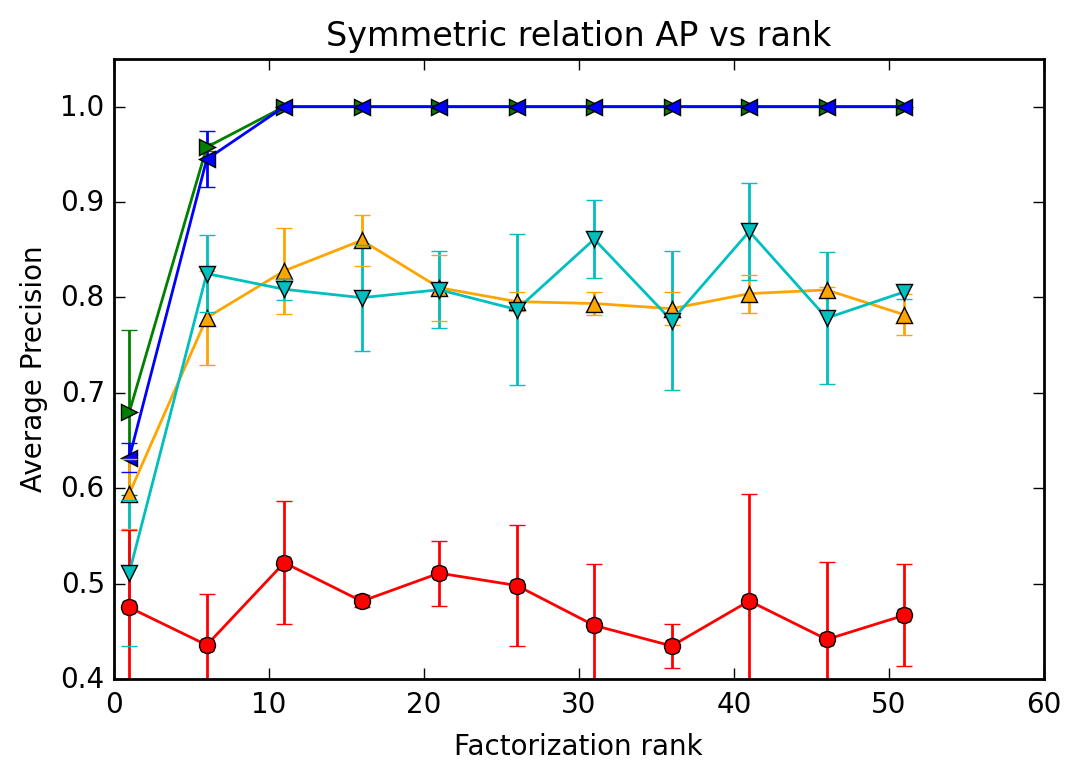}
	\includegraphics[width=0.40\textwidth]{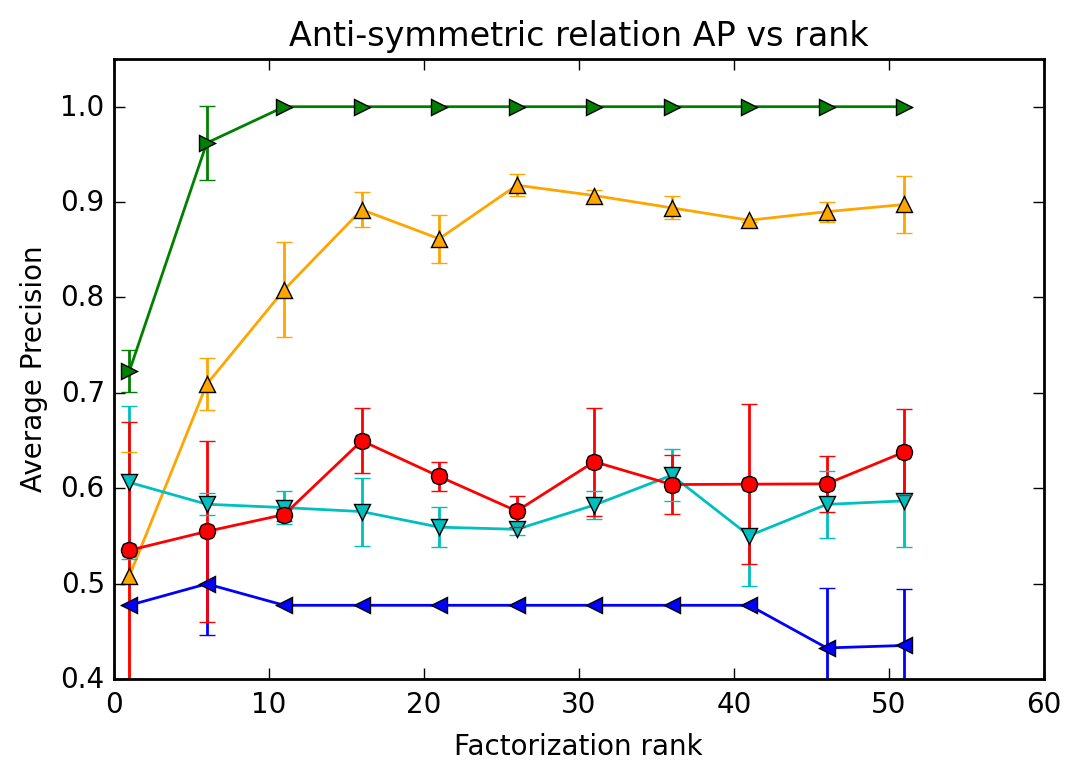} 
	\includegraphics[width=0.51\textwidth]{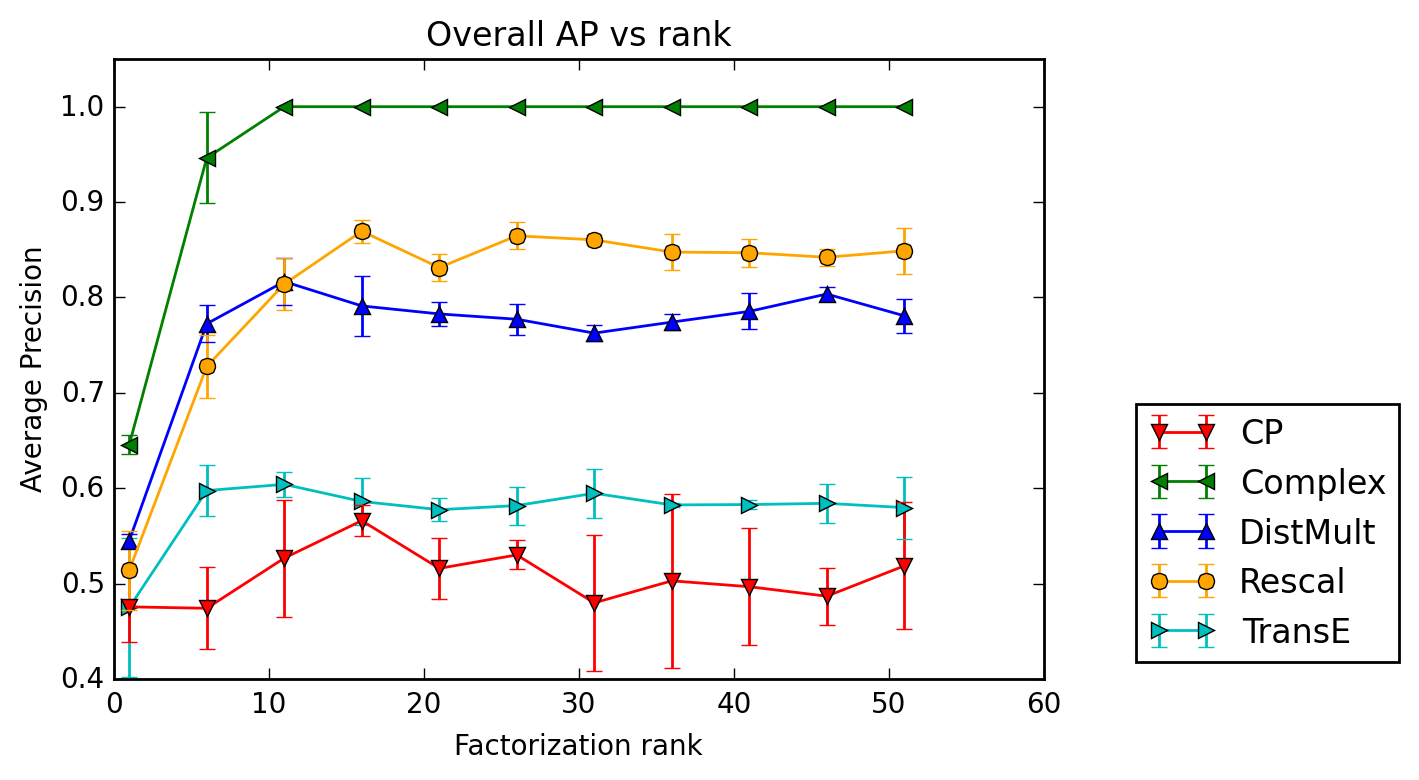}
	\caption{Average Precision (AP) for each factorization rank ranging from 1 to 50 for different state of the art models on the combined symmetry and antisymmetry experiment. Top-left: AP for the symmetric relation only. Top-right: AP for the antisymmetric relation only. Bottom: Overall AP.}
	\label{fig:exp_sym_antisym}
\end{figure*}

\subsection{Synthetic Task}

\begin{figure}
	\centering
	\includegraphics[width=0.32\linewidth]{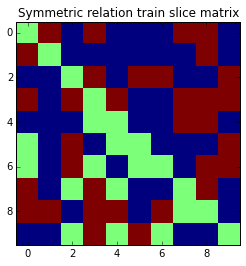}
	\includegraphics[width=0.32\linewidth]{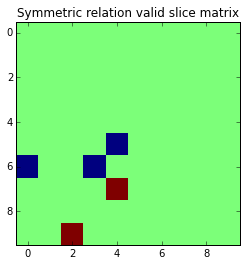}
	\includegraphics[width=0.32\linewidth]{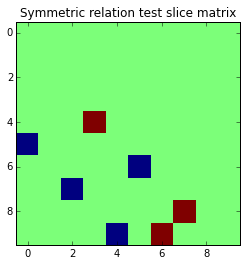}
	
	\includegraphics[width=0.32\linewidth]{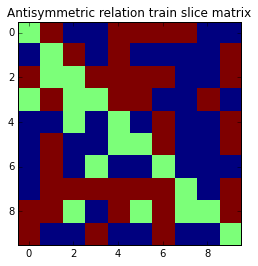}
	\includegraphics[width=0.32\linewidth]{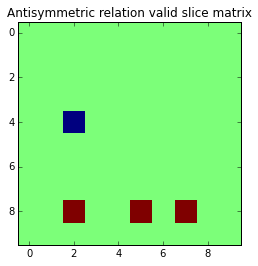}
	\includegraphics[width=0.32\linewidth]{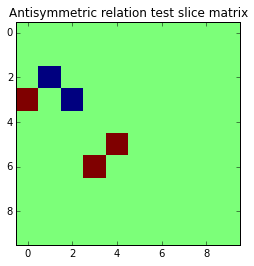}

    \caption{Parts of the training, validation and test sets of the generated experiment with one symmetric and one antisymmetric relation. Red pixels are positive triples, blue are negatives, and green missing ones. Top: Plots of the symmetric slice (relation) for the 10 first entities. Bottom: Plots of the antisymmetric slice for the 10 first entities.}
	\label{fig:symmetry_example}
\end{figure}

To assess the ability of our proposal to accurately model symmetry and antisymmetry, we randomly generated a KB of two relations and 30 entities. One relation is entirely symmetric, while the other is completely antisymmetric. This dataset corresponds to a $2 \times 30 \times 30$ tensor. Figure \ref{fig:symmetry_example} shows a part of this randomly generated tensor, with a symmetric slice and an antisymmetric slice, decomposed into training, validation and test sets. The diagonal is unobserved as it is not relevant in this experiment.

The train set contains 1392 observed triples, whereas the validation and test sets contain 174 triples each. Figure \ref{fig:exp_sym_antisym} shows the best cross-validated Average Precision (area under Precision-Recall curve) for different factorization models of ranks ranging up to 50. Models were trained using Stochastic Gradient Descent with mini-batches and AdaGrad for tuning the learning rate~\cite{duchi2011adaptive}, by minimizing the negative log-likelihood of the logistic model with $L^2$ regularization on the parameters $\Theta$ of the considered model:
\begin{equation}
    \min_{\Theta} \sum_{r(s,o) \in \Omega} \log( 1 + \exp(-\ObsTensor_{rso}\phi(s,r,o;\Theta))) + \lambda ||\Theta||^2_2\enspace.
\end{equation}
In our model, $\Theta$ corresponds to the embeddings $e_s,w_r,e_o \in \C^K$.
We describe the full algorithm in Appendix \ref{app:sgd}.

$\lambda$ is validated in $\{0.1, 0.03,$ $ 0.01, 0.003,$ $ 0.001, 0.0003,$ $ 0.00001, 0.0\}$. As expected, DistMult \cite{Yang2015} is not able to model antisymmetry and only predicts the symmetric relations correctly. Although TransE \cite{bordes2013translating} is not a symmetric model, it performs poorly in practice, particularly on the antisymmetric relation. 
RESCAL \cite{Nickel2011}, with its large number of parameters, quickly overfits as the rank grows. Canonical Polyadic (CP) decomposition \cite{hitchcock-sum-1927} fails on both relations as it has to push symmetric and antisymmetric patterns through the entity embeddings. Surprisingly, only our model succeeds on such simple data.

\begin{table*}[t]
    \centering
    \begin{tabular}{@{\extracolsep{8pt}}lllllllllll@{}}
        \toprule
        
         & \multicolumn{5}{c}{\textbf{WN18}} & \multicolumn{5}{c}{\textbf{FB15K}} \\ \cline{2-6} \cline{7-11}
         & \multicolumn{2}{c}{MRR} & \multicolumn{3}{c}{Hits at} & \multicolumn{2}{c}{MRR} & \multicolumn{3}{c}{Hits at} \\ \cline{2-3} \cline{4-6} \cline{7-8} \cline{9-11}
        
        Model & Filter & Raw & 1 & 3 & 10 & Filter & Raw & 1 & 3 & 10 \\ \hline
        
        CP & 0.075 & 0.058 & 0.049 & 0.080 & 0.125 & 0.326 & 0.152 & 0.219 & 0.376 & 0.532 \\
        TransE & 0.454 & 0.335 & 0.089 & 0.823 & 0.934 & 0.380 & 0.221 & 0.231 & 0.472 &  0.641 \\
        DistMult & 0.822 & 0.532 & 0.728 & 0.914 & 0.936 & 0.654 & \textbf{0.242} & 0.546 & 0.733   & 0.824 \\ 
        HolE* & 0.938 & \textbf{0.616} & 0.93 & \textbf{0.945} & \textbf{0.949} & 0.524 & 0.232 & 0.402 & 0.613 & 0.739\\ \hline
        ComplEx & \textbf{0.941} &  0.587 &  \textbf{0.936} &  \textbf{0.945} &  0.947 & \textbf{0.692} & \textbf{0.242} & \textbf{0.599} & \textbf{0.759}   & \textbf{0.840} \\
        
        \bottomrule
    \end{tabular}
    \caption{Filtered and Raw Mean Reciprocal Rank (MRR) for the models tested on the FB15K and WN18 datasets. Hits@m metrics are filtered. *Results reported from \cite{nickel_2016_holographic} for HolE model.}
    \label{tab:fb15k_wn18_res}
\end{table*}

\subsection{Datasets: FB15K and WN18}

\begin{table}[h]
    \centering
    \begin{tabular}{l|lll}
        Dataset & $|\EntitySpace|$ & $|\RelationSpace|$ & \#triples in Train/Valid/Test \\ \hline
        WN18 & 40,943 & 18 & 141,442 / 5,000 / 5,000 \\
        FB15K & 14,951 & 1,345 & 483,142 / 50,000 / 59,071 \\
    \end{tabular}
    \caption{Number of entities, relations, and observed triples in each split for the FB15K and WN18 datasets.}
    \label{tab:fb15k_wn18_meta}
\end{table}

We next evaluate the performance of our model on the FB15K and WN18 datasets. FB15K is a subset of \emph{Freebase}, a curated KB of general facts, whereas WN18 is a subset of \emph{Wordnet}, a database featuring lexical relations between words.  We use original training, validation and test set splits as provided by \citet{bordes2013translating}. Table \ref{tab:fb15k_wn18_meta} summarizes the metadata of the two datasets.

Both datasets contain only positive triples. As in \citet{bordes2013translating}, we generated negatives using the \emph{local closed world assumption}. That is, for a triple, we randomly change either the subject or the object at random, to form a negative example. 
 This negative sampling is performed at runtime for each batch of training positive examples.

For evaluation, we measure the quality of the ranking of each test triple among all possible subject and object substitutions
: $r(s',o)$ and $r(s,o')$, $\forall s', \forall o' \in \setent$.
 Mean Reciprocal Rank (MRR) and Hits at $m$ are the standard evaluation measures for these datasets and come in two flavours: raw and filtered \cite{bordes2013translating}. The filtered metrics are computed \emph{after} removing all the other positive observed triples that appear in either training, validation or test set from the ranking, whereas the raw metrics do not remove these. 

Since ranking measures are used, previous studies generally preferred a pairwise ranking loss for the task \cite{bordes2013translating,nickel_2016_holographic}. We chose to use the negative log-likelihood of the logistic model, as it is a continuous surrogate of the sign-rank, and has been shown to learn compact representations for several important relations, especially for transitive relations~\cite{bouchard2015}. In preliminary work, we tried both losses, and indeed the log-likelihood yielded better results than the ranking loss (except with TransE), especially on FB15K.

We report both filtered and raw MRR, and filtered Hits at 1, 3 and 10 in Table \ref{tab:fb15k_wn18_res} for the evaluated models. Furthermore, we chose TransE, DistMult and HolE as baselines since they are the best performing models on those datasets to the best of our knowledge \cite{nickel_2016_holographic,Yang2015}. We also compare with the CP model to emphasize empirically the importance of learning unique embeddings for entities. 
For experimental fairness, we reimplemented these methods within the same framework as the ComplEx model, using theano~\cite{theano}. 
However, due to time constraints and the complexity of an efficient implementation of HolE, we record the original results for HolE as reported in \citet{nickel_2016_holographic}.

\subsection{Results}

WN18 describes lexical and semantic hierarchies between concepts and contains many antisymmetric relations such as hypernymy, hyponymy, or being "part of". Indeed, the DistMult and TransE models are outperformed here by ComplEx and HolE, which are on par with respective filtered MRR scores of 0.941 and 0.938. 
Table \ref{tab:wn18_detailed_res} shows the filtered test set MRR for the models considered and each relation of WN18, confirming the advantage of our model on antisymmetric relations while losing nothing on the others. 2D projections of the relation embeddings provided in Appendix \ref{app:wn18_pca} visually corroborate the results.

\begin{table}
    \begin{tabular}{l|l@{\hspace{0.5em}}l@{}@{\hspace{0.5em}}l@{}}
        
        Relation name & ComplEx & DistMult & TransE\\ \hline
        hypernym  & \textbf{0.953} & 0.791 & 0.446 \\
        hyponym  & \textbf{0.946} & 0.710 & 0.361 \\
        member\_meronym  & \textbf{0.921} & 0.704 & 0.418 \\
        member\_holonym  & \textbf{0.946} & 0.740 & 0.465 \\
        instance\_hypernym  & \textbf{0.965} & 0.943 & 0.961 \\
        instance\_hyponym  & \textbf{0.945} & 0.940 & 0.745 \\
        has\_part  & \textbf{0.933} & 0.753 & 0.426 \\
        part\_of  & \textbf{0.940} & 0.867 & 0.455 \\
        member\_of\_domain\_topic  & \textbf{0.924} & 0.914 & 0.861 \\
        synset\_domain\_topic\_of  & \textbf{0.930} & 0.919 & 0.917 \\
        member\_of\_domain\_usage  & \textbf{0.917} & \textbf{0.917} & 0.875 \\
        synset\_domain\_usage\_of  & \textbf{1.000} & \textbf{1.000} & \textbf{1.000} \\
        member\_of\_domain\_region  & \textbf{0.865} & 0.635 & \textbf{0.865} \\
        synset\_domain\_region\_of  & 0.919 & 0.888 & \textbf{0.986} \\
        derivationally\_related\_form  & \textbf{0.946} & 0.940 & 0.384 \\
        similar\_to  & \textbf{1.000} & \textbf{1.000} & 0.244 \\
        verb\_group  & \textbf{0.936} & 0.897 & 0.323 \\
        also\_see  & 0.603 & \textbf{0.607} & 0.279 \\

    \end{tabular}
    \caption{Filtered Mean Reciprocal Rank (MRR) for the models tested on each relation of the Wordnet dataset (WN18).}
    \label{tab:wn18_detailed_res}
    
    \vspace{-5mm}
\end{table}

On FB15K, the gap is much more pronounced and the ComplEx model largely outperforms HolE, with a filtered MRR of 0.692 and 59.9\% of Hits at 1, compared to 0.524 and 40.2\% for HolE. We attribute this to the simplicity of our model and the different loss function. This is supported by the relatively small gap in MRR compared to DistMult (0.654); our model can in fact be interpreted as a complex number version of DistMult. 
On both datasets, TransE and CP are largely left behind. This illustrates the power of the simple dot product in the first case, and the importance of learning unique entity embeddings in the second. CP performs poorly on WN18 due to the small number of \mbox{relations}, which magnifies this subject/object difference.

Reported results are given for the best set of hyper-parameters evaluated on the validation set for each model, after grid search on the following values: $\rank \in \{10,20,50,100,150,200\}$, $\lambda \in \{0.1, 0.03, 0.01, 0.003, 0.001, 0.0003,0.0\}$, $\alpha_0 \in \{1.0, 0.5, 0.2, 0.1, 0.05, 0.02, 0.01\}$,  $\eta \in \{1, 2, 5, 10\}$ with $\lambda$ the $L^2$ regularization parameter, $\alpha_0$ the initial learning rate (then tuned at runtime with AdaGrad), and $\eta$ the number of negatives generated per positive training triple. We also tried varying the batch size but this had no impact and we settled with 100 batches per epoch. Best ranks were generally 150 or 200, in both cases scores were always very close for all models. The number of negative samples per positive sample also had a large influence on the filtered MRR on FB15K (up to +0.08 improvement from 1 to 10 negatives), but not much on WN18. On both datasets regularization was important (up to +0.05 on filtered MRR between $\lambda=0$ and optimal one).
We found the initial learning rate to be very important on FB15K, while not so much on WN18. We think this may also explain the large gap of improvement our model provides on this dataset compared to previously published results -- as DistMult results are also better than those previously reported \cite{Yang2015} -- along with the use of the log-likelihood objective. It seems that in general AdaGrad is relatively insensitive to the initial learning rate, perhaps causing some overconfidence in its ability to tune the step size online and consequently leading to less efforts when selecting the initial step size.

Training was stopped using early stopping on the validation set filtered MRR, computed every 50 epochs with a maximum of 1000 epochs.

\subsection{Influence of Negative Samples}
We further investigated the influence of the number of negatives generated per positive training sample. In the previous experiment, due to computational limitations, the number of negatives per training sample, $\eta$, was validated among the possible numbers $\{1, 2, 5, 10\}$. We want to explore here whether increasing these numbers could lead to better results. To do so, we focused on FB15K, with the best validated $\lambda,\rank,\alpha_0$, obtained from the previous experiment. We then let $\eta$ vary in $\{1, 2, 5, 10, 20, 50, 100, 200\}$.

\begin{figure}[h]
	\centering
	\includegraphics[width=0.99\linewidth]{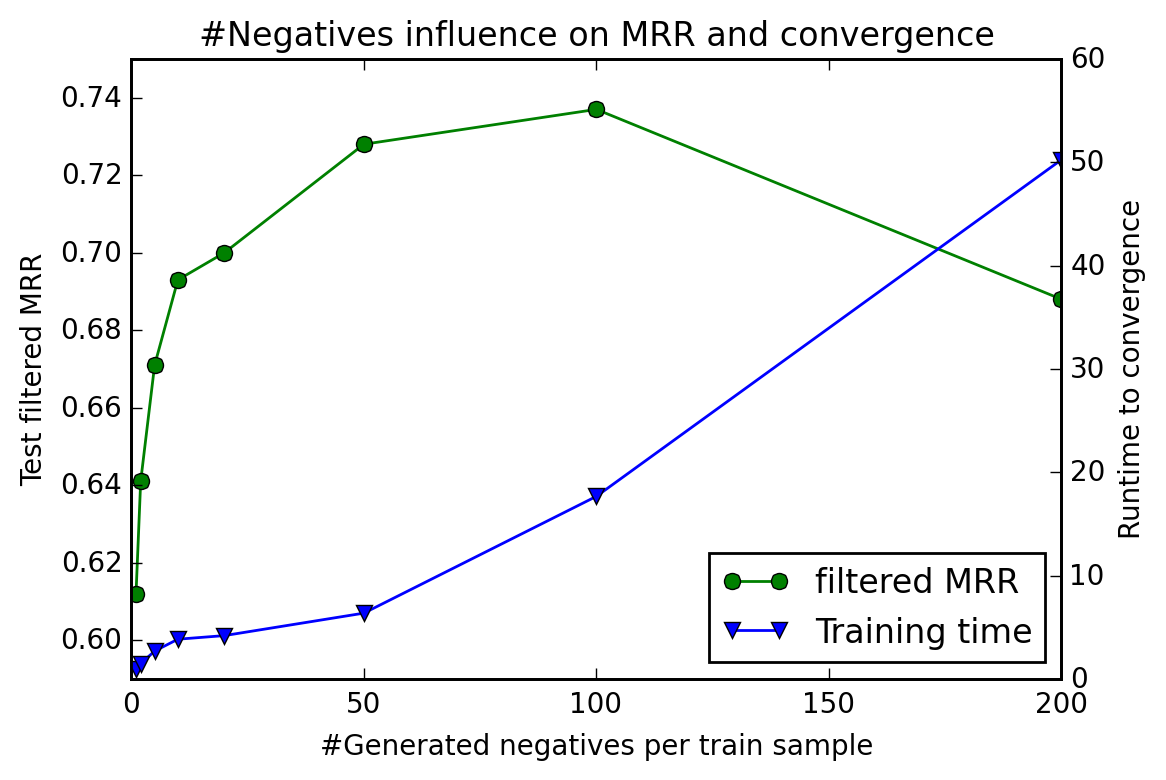}
	
    \caption{Influence of the number of negative triples generated per positive training example on the filtered test MRR and on training time to convergence on FB15K for the ComplEx model with $\rank=200$, $\lambda=0.01$ and $\alpha_0=0.5$. Times are given relative to the training time with one negative triple generated per positive training sample ($=1$ on time scale).}
	\label{fig:neg_ratio}
	
\end{figure}

Figure \ref{fig:neg_ratio} shows the influence of the number of generated negatives per positive training triple on the performance of our model on FB15K.
Generating more negatives clearly improves the results, with a filtered MRR of 0.737  with 100 negative triples (and 64.8\% of Hits@1), before decreasing again with 200 negatives. The model also converges with fewer epochs, which compensates partially for the additional training time per epoch, up to 50 negatives. It then grows linearly as the number of negatives increases, making 50 a good trade-off between accuracy and training time. 

\section{Related Work}
In the early age of spectral theory in linear algebra, complex numbers were not used 
for matrix factorization and mathematicians mostly focused on bi-linear forms~\cite{beltrami1873sulle}. 
The eigen-decomposition in the complex domain as taught today in linear algebra courses 
came 40 years later~\cite{autonne1915}. Similarly, most of the existing approaches for tensor 
factorization were based on decompositions in the real domain, such as the Canonical Polyadic (CP) decomposition~\cite{hitchcock-sum-1927}. These methods are very effective in many applications 
that use different modes of the tensor for different types of entities.
But in the link prediction problem, antisymmetry of relations was quickly seen as 
a problem and asymmetric extensions of tensors were studied, mostly by either
considering independent embeddings~\cite{sutskever2009} or considering relations as
matrices instead of vectors in the RESCAL model~\cite{Nickel2011}. 
Direct extensions were based on uni-,bi- and trigram latent factors for triple data, as well as a low-rank relation matrix~\cite{Jenatton2012}. 

Pairwise interaction models were also considered to improve prediction performances. For example, the Universal Schema approach~\cite{riedel_2013_univschema} factorizes a 2D unfolding of the tensor (a matrix of entity pairs vs. relations) while \citet{Welbl2016} extend this also to other pairs.

In the Neural Tensor Network (NTN) model, \citet{socher2013reasoning} combine linear transformations and multiple bilinear forms of subject and object embeddings to jointly feed them into a nonlinear neural layer. Its non-linearity and multiple ways of including interactions between embeddings gives it an advantage in expressiveness over models with simpler scoring function like DistMult or RESCAL. As a downside, its very large number of parameters can make the NTN model harder to train and overfit more easily.

The original multi-linear DistMult model is symmetric in subject and object for every relation ~\cite{Yang2015} and achieves good performance, presumably due to its simplicity.
The TransE model from \citet{bordes2013translating} also embeds entities and relations in the same space and imposes a geometrical structural bias into the model: the subject entity vector should be close to the object entity vector once translated by the relation vector.

A recent novel way to handle antisymmetry is via the Holographic Embeddings (HolE) model by ~\cite{nickel_2016_holographic}. In HolE the circular correlation is used for combining entity embeddings, measuring the covariance between embeddings at different dimension shifts. This generally suggests that other composition functions than the classical tensor product can be  helpful as they allow for a richer interaction of embeddings. However, the asymmetry in the composition function in HolE stems from the asymmetry of circular correlation, an $\mathcal{O}(n log(n))$ operation, whereas ours is inherited from the complex inner product, in $\mathcal{O}(n)$.

\section{Conclusion}

We described a simple approach to matrix and tensor factorization for link prediction data that uses vectors with complex values and retains the mathematical definition of the dot product. 
The class of normal matrices is a natural fit for binary relations, and using the real part allows for efficient approximation of any learnable relation. Results on standard benchmarks show that no more modifications are needed to improve over the state-of-the-art.


There are several directions in which this work can be extended. An obvious one is to merge our approach with known extensions to tensor factorization in order to further improve predictive performance. For example, the use of pairwise embeddings together with complex numbers might lead to improved results in many situations that involve non-compositionality. 
Another direction would be to develop a more intelligent negative sampling procedure, to generate more informative negatives with respect to the positive sample from which they have been sampled. It would reduce the number of negatives required to reach good performance, thus accelerating training time.

Also, if we were to use complex embeddings every time a model includes a dot product, e.g. in deep neural networks, would it lead to a similar systematic improvement? 



\section*{Acknowledgements} 
This work was supported in part by the Paul
Allen Foundation through an Allen Distinguished
Investigator grant and in part by a Google Focused Research Award.

\bibliography{complex_bib,nonauto_bib}
\bibliographystyle{icml2016}

\clearpage




\appendix

\section{SGD algorithm}
\label{app:sgd}
We describe the algorithm to learn the ComplEx model with Stochastic Gradient Descent using only real-valued vectors.

Let us rewrite equation \ref{eqn:sesquilinear-dot}, by denoting
the real part of embeddings with primes and the imaginary part with double primes: 
$e'_i = \real(e_i)$, $e''_i = \imag(e_i)$, $w'_r = \real(w_r)$, $w''_r = \imag(w_r)$.
The set of parameters is $\Theta=\{e'_i,e''_i,w'_r,w''_r; \forall
i \in \setent, \forall r \in \setrel \}$,
and the scoring function involves only real vectors:

\begin{eqnarray}
\phi(r,s,o;\Theta) &=& \left<w'_r,e'_s, e'_o\right>+ \left<w'_r,e''_s, e''_o\right> \notag\\
    &&+ \left<w''_r, e'_s, e''_o\right> - \left<w''_r,e''_s, e'_o\right>\notag
\end{eqnarray}

where each entity and each relation has two real embeddings.

Gradients are now easy to write:

\begin{eqnarray}
\label{gradients}
\grad_{e'_s} \phi(r,s,o;\Theta) &=& (w'_r \odot e'_o) + (w''_r \odot e''_o)\notag\\
\grad_{e''_s} \phi(r,s,o;\Theta) &=& (w'_r \odot e''_o) - (w''_r \odot e'_o)\notag\\
\grad_{e'_o} \phi(r,s,o;\Theta) &=& (w'_r \odot e'_s) - (w''_r \odot e''_s)\notag\\
\grad_{e''_o} \phi(r,s,o;\Theta) &=& (w'_r \odot e''_s) + (w''_r \odot e'_s)\notag\\
\grad_{w'_r} \phi(r,s,o;\Theta) &=& (e'_s \odot e'_o) + (e''_s \odot e''_o)\notag\\
\grad_{w''_r} \phi(r,s,o;\Theta) &=& (e'_s \odot e''_o) - (e''_s \odot e'_o)\notag
\end{eqnarray}

where $\odot$ is the element-wise (Hadamard) product.

As stated in equation \ref{observation-model} we use the sigmoid link function, and minimize the $L^2$-regularized negative log-likelihood:

\begin{eqnarray}
    \gamma(\Omega;\Theta) &=&
    \sum_{r(s,o) \in \Omega} \log( 1 + \exp(-\ObsTensor_{rso}\phi(s,r,o;\Theta)))\notag\\
    &&+ \lambda ||\Theta||^2_2\enspace.\notag
\end{eqnarray}

To handle regularization, note that the squared $L^2$-norm of a complex vector $v=v'+iv''$
is the sum of the squared modulus of each entry:

\begin{eqnarray}
||v||^2_2 &=& \sum_j \sqrt{v_j^{\prime2} + v_j^{\prime\prime2}}^2\notag\\
&=& \sum_j v_j^{\prime2} + \sum_j  v_j^{\prime\prime2}\notag\\
&=& ||v'||^2_2 +  ||v''||^2_2\notag
\end{eqnarray}

which is actually the sum of the $L^2$-norms of the vectors of the real and imaginary parts.

We can finally write the gradient of $\gamma$ with respect to a \emph{real} embedding $v$ for
one triple $r(s,o)$:

\begin{eqnarray}
    \grad_v \gamma(\{r(s,o)\};\Theta) &=& -\ObsTensor_{rso}\phi(s,r,o;\Theta) \sigma(\grad_v \phi(r,s,o;\Theta))\notag\\
    &&+ 2\lambda v\notag
\end{eqnarray}

where $\sigma(x) = \frac{1}{1+\mathrm{e}^{-x}}$ is the sigmoid function.

Algorithm \ref{SGDC} describes SGD for this formulation of the
scoring function.
When $\Omega$ contains only positive triples,
we generate $\eta$ negatives per positive train triple, by corrupting
either the subject or the object of the positive triple, as 
described in \citet{bordes2013translating}.

\begin{algorithm}[t]
\caption{SGD for the ComplEx model}
\label{SGDC}
\begin{algorithmic}
\INPUT Training set $\Omega$, Validation set $\Omega_v$, learning rate $\alpha$, embedding dim. $k$, regularization factor $\lambda$, negative ratio $\eta$, batch size $b$, max iter $m$, early stopping $s$.
\STATE $e'_i \gets \randn(k)$, $e''_i \gets \randn(k)$ for each $i \in \mathcal{E}$
\STATE $w'_i \gets \randn(k)$, $w''_i \gets \randn(k)$ for each $i \in \mathcal{R}$
\FOR{$i=1,\cdots,m$}
    \FOR {$j=1..|\Omega|/b$}
        \STATE $\Omega_b \gets$ sample$(\Omega,b,\eta)$
        \STATE Update embeddings w.r.t.:\\
        $\quad\quad\sum_{r(s,o) \in \Omega_b} \grad \gamma(\{r(s,o)\};\Theta)$
        \STATE Update learning rate $\alpha$ using Adagrad
    \ENDFOR
    \IF{$i \mod s = 0$}
        \STATE \textbf{break} if filteredMRR or AP on $\Omega_v$ decreased
    \ENDIF
\ENDFOR
\end{algorithmic}
\end{algorithm}

\section{WN18 embeddings visualization}
\label{app:wn18_pca}

\begin{figure*}[!ht]
	\centering
	\includegraphics[width=0.49\textwidth]{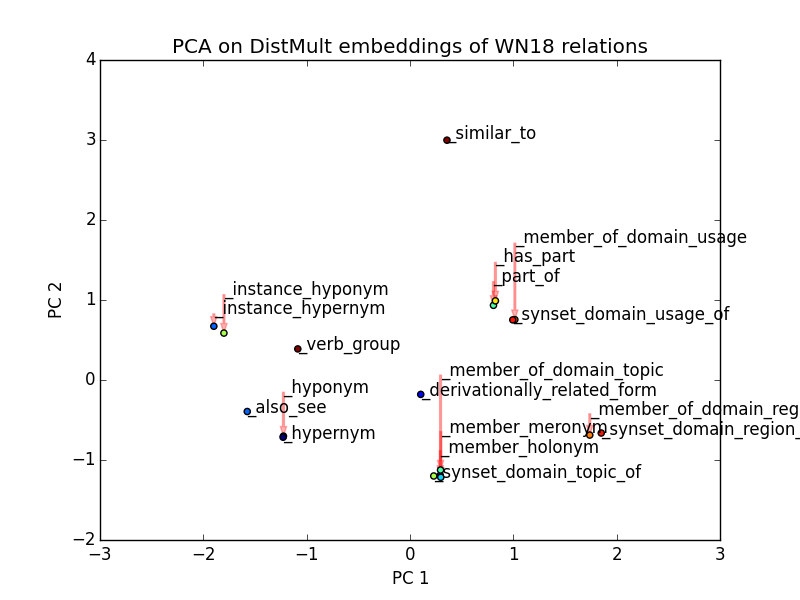}
	\includegraphics[width=0.49\textwidth]{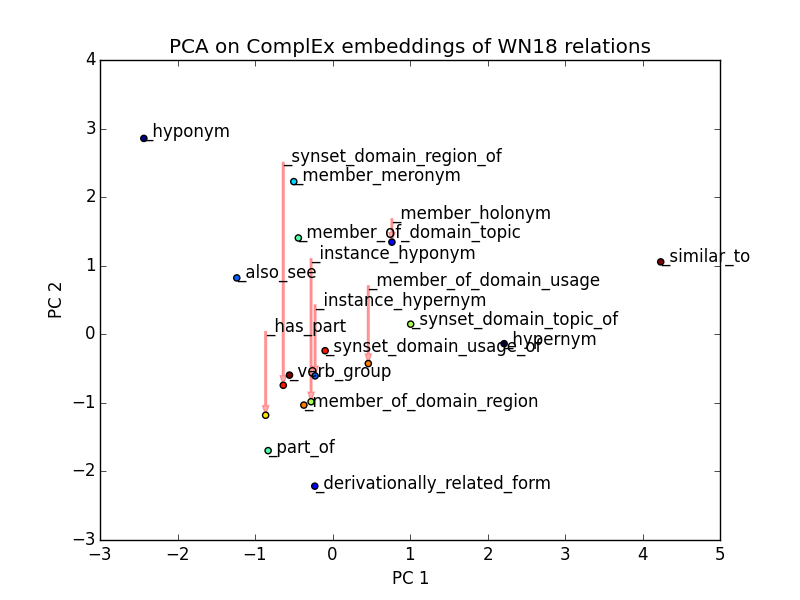}
	\includegraphics[width=0.49\textwidth]{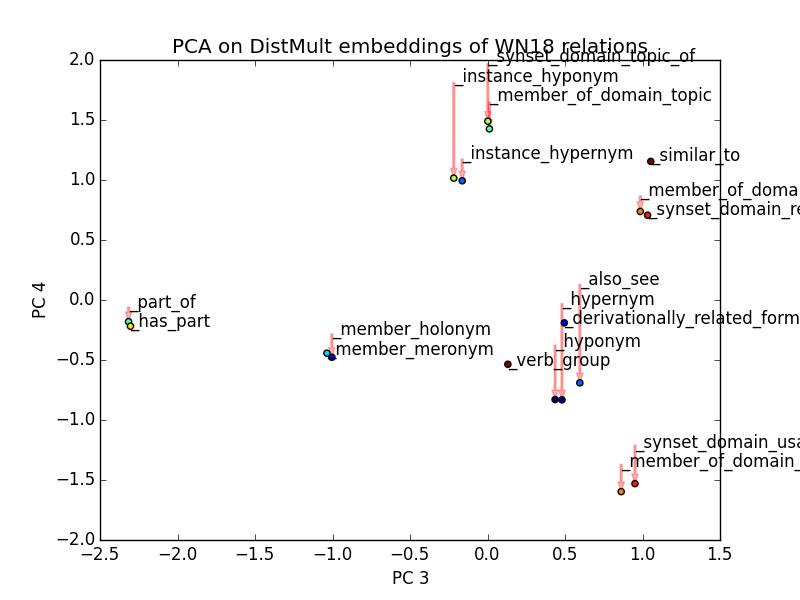}
	\includegraphics[width=0.49\textwidth]{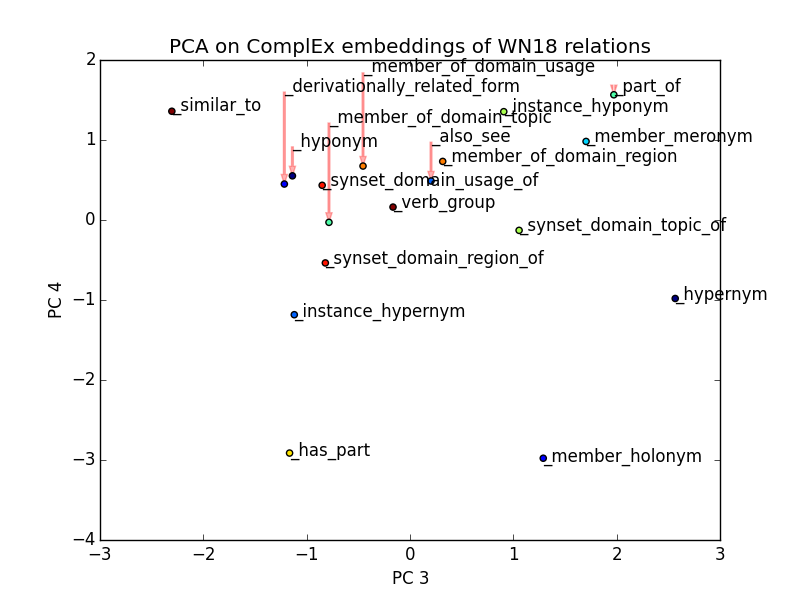}
	\vspace{-5mm}
	\caption{Plots of the first and second (Top), third and fourth (Bottom) components of the WN18 relations embeddings using PCA. Left: DistMult embeddings. Right: ComplEx embeddings. Opposite
	relations are clustered together by DistMult while correctly separated by ComplEx.}
	\label{fig:pca}
\end{figure*}

We used principal component analysis (PCA) to visualize embeddings of the relations 
of the wordnet dataset (WN18). We plotted the four first components of the best DistMult
and ComplEx model's embeddings in Figure \ref{fig:pca}. For the ComplEx model, we simply concatenated
the real and imaginary parts of each embedding. 

Most of WN18 relations describe hierarchies, and are thus antisymmetric.
Each of these hierarchic relations has its inverse relation in the dataset. For example: \texttt{hypernym} / \texttt{hyponym},
\texttt{part\_of} / \texttt{has\_part}, \texttt{synset\_domain\_topic\_of} / \texttt{member\_of\_domain\_topic}.
Since DistMult is unable to model antisymmetry, it will correctly represent the nature
of each pair of opposite relations, but not the direction of the relations.
Loosely speaking, in the \texttt{hypernym} / \texttt{hyponym} pair the nature is 
sharing semantics,
and the direction is that one entity generalizes the semantics of the other. 
This makes DistMult reprensenting the opposite relations with very close embeddings,
as Figure \ref{fig:pca} shows. It is especially striking for the third and
fourth principal component (bottom-left). Conversely, ComplEx manages to oppose spatially
the opposite relations.

\end{document}